\documentclass[lettersize,journal]{IEEEtran}
\usepackage{amsmath,amsfonts}
\usepackage{algorithmic}
\usepackage{algorithm}
\usepackage{array}
\usepackage[caption=false,font=normalsize,labelfont=sf,textfont=sf]{subfig}
\usepackage{textcomp}
\usepackage{stfloats}
\usepackage{url}
\usepackage{verbatim}
\usepackage{graphicx}
\usepackage{cite}
\usepackage{multirow}
\usepackage{colortbl}
\usepackage{xcolor}
\usepackage{booktabs}
\usepackage{pifont}        

\usepackage{booktabs}
\usepackage{tabularx}
\usepackage{caption}
\usepackage{siunitx}
\usepackage{makecell}

\RequirePackage{xspace}
\makeatletter
\DeclareRobustCommand\onedot{\futurelet\@let@token\@onedot}
\def\@onedot{\ifx\@let@token.\else.\null\fi\xspace}

\makeatother

\definecolor{rblue}{rgb}{0,0.5,1}
\definecolor{awesome}{rgb}{1.0, 0.13, 0.32}
\definecolor{hollywoodcerise}{rgb}{0.96, 0.0, 0.63}
\definecolor{lasallegreen}{rgb}{0.03, 0.47, 0.19}
\definecolor{hanpurple}{rgb}{0.32, 0.09, 0.98}
\definecolor{green(pigment)}{rgb}{0.0, 0.65, 0.31}

\usepackage[most]{tcolorbox} 

\newtcolorbox{bluequestion}{
  enhanced, breakable,
  colback=blue!5,        
  colframe=blue!30,      
  boxrule=0pt,           
  borderline west={6pt}{0pt}{blue!60}, 
  left=8pt, right=8pt, top=6pt, bottom=6pt, 
  arc=0pt                
}

\definecolor{cvprblue}{rgb}{0.21,0.49,0.74}
\definecolor{DeepPink}{rgb}{1,0.078,0.576}
\usepackage[pagebackref,breaklinks,colorlinks,allcolors=DeepPink]{hyperref}

\begin{document}

\title{GeoTLM: Geometry-aware Tactile-Language Models for Contact Motion Orientation Reasoning of Dynamic Objects}

\author{Qiutian Li, Zinan Liu, and Lin Wang$^{\dag}$~\IEEEmembership{Member,~IEEE}
\thanks{$^{\dag}$ Corresponding author.}
\thanks{Qiutian Li, Zinan Liu and Lin Wang are with the School of EEE, Nanyang Technological University (NTU), Singapore (email: qiutian001@e.ntu.edu.sg; zinan001@e.ntu.edu.sg; linwang@ntu.edu.sg). }}

\markboth{Journal of \LaTeX\ Class Files,~Vol.~14, No.~8, August~2021}%
{Shell \MakeLowercase{\textit{et al.}}: A Sample Article Using IEEEtran.cls for IEEE Journals}


\maketitle

\begin{abstract}
Modern tactile-language models (TLMs) have shown potential for  robot learning tasks, such as  material and texture recognition.
However, for contact-rich scenarios, these TLMs struggle to understand the physical properties of dynamic objects, such as rotation and sliding directions. For instance, our preliminary experiments reveal that popular TLMs, such as Sparsh and AnyTouch2, exhibit weak performance on basic rotation direction reasoning from GelSight Mini tactile data. 

This surprising gap inspires us to explore a \textbf{novel research question}: \textit{Can we inject physically grounded geometric priors into TLMs to enable reliable contact orientation reasoning of dynamic object properties?} 
To this end, we propose \textbf{GeoTLM}, a novel geometric representation-guided TLM for the perception of dynamic contact events. 
\textbf{Our key idea} is \textit{to preserve and structure tactile shear-field geometry before language-level reasoning, rather than forcing low-resolution tactile tokens into fragile closed-form physics operators}. To achieve this, we propose a lightweight (only 14k parameters) yet novel \textbf{Differentiable Geometric Representation} (DGR).
Specifically, DGR learns a contact-mask-guided representation in the shear field and aggregates it through an antisymmetric seven-region pooling design, motivated by the physical intuition that rotational contact produces antisymmetric deformation patterns.
We conduct experiments on two representative tasks: rotation direction and sliding direction reasoning.
Extensive experiments show that GeoTLM improves novel-object rotation accuracy by \textbf{+14.6\%} and real-sensor sliding accuracy by \textbf{+16.2\%} over the same backbone without the geometric encoder.
Obviously, GeoTLM benefits from preserving raw geometric shear structures for language-level recombination.
Overall, our work paves a new  way for physically grounded tactile-language reasoning, with strong potential for dynamic object understanding and contact-rich robotic manipulation.

\end{abstract}




\section{Introduction}
\label{sec:intro}

Modern tactile-language models (TLMs)~\cite{yu2024octopi, xie2026universal, cheng2026stola} hold strong potential for robotic manipulation by bridging contact-rich tactile perception with semantic reasoning. In real-world manipulation, robots must continuously interpret dynamic physical events from tactile signals in real time. Tightening a bottle cap requires identifying rotation direction at each step, peg insertion demands early detection of incipient slip, and safe grasping relies on estimating contact force before the object is crushed. These requirements go beyond static texture or contact recognition, calling for models that can understand evolving physical properties such as rotation direction, sliding direction, and applied force. TLMs have recently emerged as a promising interface for this goal, as their language heads can connect tactile perception with downstream task reasoning, tool use, and human instructions~\cite{yu2024octopi, huang2025tactile}.

Despite this promise, current TLMs remain surprisingly limited in contact-rich dynamic reasoning.
The dominant recipe in tactile perception is to pretrain a large encoder on hundreds of thousands to millions of tactile frames and then either probe the representation or attach a language head~\cite{higuera2024sparsh, feng2026anytouch}.
This paradigm performs well under standard distribution evaluation.
However, our preliminary experiments (see Tab.~\ref{tab:rotation-compare}) expose a striking weakness on a deceptively simple physical reasoning task.
Classifying a single GelSight Mini contact~\cite{yuan2017gelsight} as clockwise or counterclockwise rotation, large tactile and vision backbones probed under a frozen protocol stay close to chance ($\sim 50\%$), even though the same backbones perform well on in-distribution recognition. The signal needed to separate the two rotation directions is present in the contact, but the pretrained representation does not expose it in a form a classifier can read.

We interpret this gap as a failure of inductive bias rather than capacity: an $87$M-parameter tactile encoder and larger vision backbones alike miss a geometric signal that a compact, structured model can exploit.
Rotational contact between an elastomer pad and a rigid object produces a characteristic antisymmetric shear pattern, and the cue for chirality lies in this local geometric structure.
Generic V-JEPA and contrastive pretraining objectives are not designed to preserve such antisymmetric shear at the patch-grid resolution used by downstream tokens.
Therefore, adding more frames or more parameters does not necessarily recover physical information that has already been compressed away.

This observation motivates our central research question: \textit{Can physically grounded geometric priors be injected into TLMs to enable reliable perception of dynamic object properties?}
However, addressing this question is non-trivial for several reasons.
\textbf{{First}}, dynamic tactile properties are encoded in subtle geometric deformation, where rotation and sliding cues arise from local shear and contact-flow patterns rather than static texture. These cues can be easily suppressed when tactile frames are compressed into generic patch tokens without explicit geometric preservation~\cite{higuera2024sparsh}.
\textbf{{Second}}, directly imposing handcrafted physical operators such as curl, divergence, or strain is not enough either, as such operators can be task-specific and brittle across diverse contact patterns and sensor resolutions.
Collectively, these factors suggest that reliable TLMs should not merely scale generic encoders or append fixed physical descriptors, but should preserve geometric contact structures in a form that downstream reasoning can flexibly interpret and recombine.

To this end, we introduce \textbf{{GeoTLM}}, a \textit{geometric representation-guided tactile-language model} for dynamic contact event perception.
\textbf{Our key idea} is to preserve tactile shear-field geometry before downstream reasoning, rather than forcing low-resolution tactile tokens into fragile closed-form physical operators.
GeoTLM achieves this through a lightweight \textbf{Differentiable Geometric Representation} (DGR) module (Sec.~\ref{sec:method}) with only $14$K trainable parameters, which learns contact-mask-guided shear representations from frozen tactile patch tokens and aggregates them via antisymmetric seven-region pooling.
This design is motivated by the physical intuition that rotational contact produces opposite-sign deformation patterns across spatial regions.

In summary, {\textbf{our main contributions}} are:
({\textbf{I}}) We present, to the best of our knowledge, the \textbf{first} \textit{geometry-aware TLM paradigm} \textbf{GeoTLM} for dynamic contact event perception, establishing a new direction for physically grounded tactile-language reasoning in contact-rich robotic manipulation.
({\textbf{II}}) We reveal a critical gap in current tactile-language models: while large tactile encoders perform well under in-distribution evaluation, they fail on basic dynamic contact reasoning, indicating that the bottleneck lies in geometric inductive bias rather than model capacity.
({\textbf{III}}) We design a lightweight \textbf{Differentiable Geometric Representation} (DGR) module with only \textbf{14k} trainable parameters, which preserves tactile shear-field geometry through contact-mask-guided shear projection and antisymmetric seven-region pooling, and show that it substantially improves novel-object generalization on real tactile sensors.

\section{Related Work}

\noindent\textbf{Tactile foundation models and tactile-language models.}
Self-supervised pretraining on large tactile corpora has produced general-purpose encoders for vision-based touch. Sparsh~\cite{higuera2024sparsh} trains a V-JEPA objective on roughly $460$K tactile frames, UniTouch~\cite{yang2024binding} binds touch to vision and language embeddings, and AnyTouch~\cite{feng2025anytouch} together with AnyTouch~2~\cite{feng2026anytouch} unify static and dynamic representations across optical tactile sensors. A related line aligns touch with language through paired captions or contrastive pretraining~\cite{fu2024touch, ma2026cltp, ma2026fg}. Built on top of such encoders, tactile-language models attach a language head for higher-level reasoning: Octopi~\cite{yu2024octopi, yu2025demonstrating} reasons over physical object properties, STOLA~\cite{cheng2026stola} targets open-ended tactile commonsense, and VTV-LLM~\cite{xie2026universal} extends video understanding to visuotactile streams. On the control side, TactileVLA~\cite{huang2025tactile}, OmniVTLA~\cite{cheng2025omnivtla}, and reactive policies~\cite{xue2025reactive} couple touch to action. These systems commonly use generic tactile perception tokens from pretrained encoders, and the contact-specific geometry that distinguishes one motion from another is never made explicit. Our work investigates whether a compact structured module can recover such geometry.
Whether a representation contains a particular piece of information and whether a simple readout can access it are distinct questions. Linear probing was introduced precisely to separate the two: a linear classifier on frozen features measures what the representation exposes without the confound of a trainable backbone~\cite{alain2016understanding}, and it has since become the standard protocol for evaluating self-supervised encoders, including in touch~\cite{higuera2024sparsh}. A parallel literature shows that high-capacity models often succeed for the wrong reasons, latching onto superficial correlations rather than the intended signal~\cite{geirhos2020shortcut, mccoy2019right}, and that constraining a model toward the right explanatory structure can correct this~\cite{ross2017right}. Our diagnostic sits in this tradition. We probe several strong tactile and vision backbones on rotation direction and find them near chance, which indicates that the discriminative cue, though physically present in the contact, is not linearly accessible from the pretrained features. The gap is one of inductive bias rather than capacity.

\noindent\textbf{Geometric inductive bias.}
A long line of work argues that encoding the symmetries of a problem directly into the model improves both data efficiency and generalization. Group-equivariant networks build invariance and equivariance to transformation groups into the architecture~\cite{cohen2016group}, and the broader geometric deep learning view recasts many successful models as instances of structuring computation around the symmetries of the domain~\cite{bronstein2021geometric}. Rotation orientation is a natural target for this perspective: clockwise and counterclockwise contact motions are related by reflection, so a representation whose pooling is antisymmetric under that reflection couples directly to the distinction, whereas a symmetric pool cancels it. Rather than imposing a fixed equivariant operator, our encoder builds the antisymmetric structure into the pooling while leaving the underlying shear features learnable, which we find transfers better than hand-specified differential operators.

\noindent\textbf{Dynamic contact signals.}
The temporal evolution of shear and surface deformation carries information about contact events well beyond static texture. Early GelSight work measured shear and detected slip directly from marker displacements~\cite{yuan2015measurement}. Learned features have been used for slip prediction and grasp stabilization~\cite{veiga2015stabilizing}, event cameras enable fast incipient-slip detection~\cite{rigi2018novel}, and recent visuotactile systems estimate in-hand rotation rate from incipient-slip cues~\cite{li2024incipient}. NormalFlow~\cite{huang2024normalflow}, whose dataset we use, recovers full six-degree-of-freedom contact pose by registering surface normals across frames, and demonstrates that fine contact motion is recoverable from a GelSight image when the geometry is modelled explicitly. Force estimation from optical tactile images makes the same point for a different quantity~\cite{shahidzadeh2025feelanyforce, helmut2025learning}. These methods typically recover one quantity at a time, often with sensor-specific calibration or task-tuned pipelines. We instead read motion direction from a frozen general-purpose encoder through a compact geometric module.

\noindent\textbf{Spatial reasoning and tactile localization.}
In the visual domain, SpatialVLM~\cite{chen2024spatialvlm} argues that spatial ability in vision-language models is more a matter of training data than architecture, and benchmarks such as 3DSRBench~\cite{ma20253dsrbench} and Spatial457~\cite{wang2025spatial457} probe object-centric 3D spatial reasoning in large multimodal models. These operate over RGB and depth, whose statistics differ sharply from an elastomer deformation image, and they do not address the tactile setting we study. Closer to manipulation, spatially anchored tactile awareness~\cite{huang2025spatially} grounds tactile features in robot coordinates for dexterous control, through policy learning rather than encoder design. A separate body of work localizes a sensor on a known object model~\cite{bauza2023tac2pose, suresh2023midastouch, zhang2026tacloc, suresh2024neuralfeels}, recovering \emph{where} the contact is.

\section{Methodology}
\label{sec:method}

\begin{figure*}[t!]
\centering
\includegraphics[width=\linewidth]{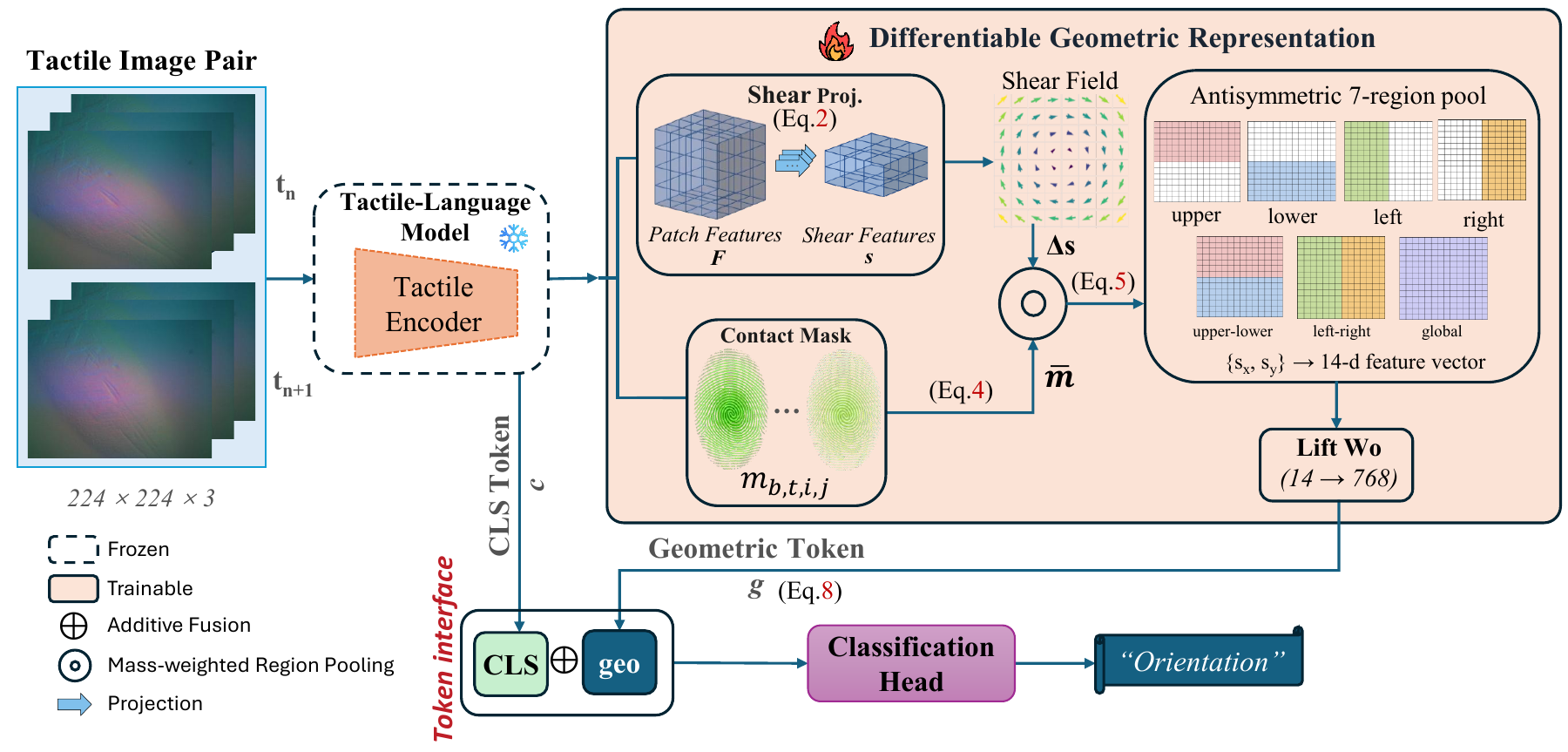}
\vspace{-16pt}
\caption{Overview of the proposed GeoTLM framework with the key design of Differentiable Geometric Representation module.}
\label{fig:arch}
\vspace{-18pt}
\end{figure*}

\subsection{Overview and Problem Definition}
\label{sec:method:overview}

Our goal is to recover the direction of contact motion from a short tactile clip, using a frozen tactile representation backbone without fine-tuning it. The difficulty is that strong tactile foundation features, while rich, do not expose motion direction in a form a linear classifier can read off, and they degrade sharply when transferred from simulated to real sensors. We address this with a small differentiable model, DGR (Differentiable Geometric Representation), that reads the patch-token stream of the frozen backbone and produces a single feature vector summarizing the directional structure of the contact deformation. This vector is added to the backbone's global token, and a linear probe is trained on the sum. The backbone stays frozen throughout; only DGR and the linear head are trained, which keeps the added parameter count to roughly $14$K. Figure~\ref{fig:arch} shows the overall structure.

The encoder is built around one observation. Rotation direction is an antisymmetric property of the contact field: a clockwise twist and its counter-clockwise mirror produce shear deformations of opposite sign across symmetric regions of the sensor. A pooling operator that is itself antisymmetric along spatial axes therefore couples directly to chirality, whereas a symmetric global pool averages the signal away. DGR makes this structure explicit while leaving the actual shear features to be learned, rather than imposing a fixed analytic operator; in our ablations the learned formulation is what makes the encoder transfer, and hand-specified differential operators degrade it.

\subsection{Shear Field and Temporal Difference}

Given a tactile clip $X \in \mathbb{R}^{B \times T \times 3 \times H \times W}$ with $T{=}4$ frames at $H{=}W{=}224$, the frozen AnyTouch~2 ViT-B/16~\cite{feng2026anytouch} applies a temporal-tube patch embedding of size two, which collapses the four input frames into $T'{=}2$ time-bins on a $14\times14$ patch grid. A single forward pass returns the token sequence. We take the global token $c \in \mathbb{R}^{768}$ for the later fusion, discard the five sensor-type tokens, and keep the 392 patch tokens as the input to DGR:
\begin{equation}
\footnotesize
\mathbf{F} = \mathrm{reshape}(\mathbf{F};\, B, T'{=}2, 14, 14, 768) \in \mathbb{R}^{B \times 2 \times 14 \times 14 \times 768},
\end{equation}
so each entry $\mathbf{F}_{b,t,i,j} \in \mathbb{R}^{768}$ is the feature of patch $(i,j)$ in the time-bin $t \in \{0,1\}$. Both $c$ and $\mathbf{F}$ come from the same frozen backbone and the same forward pass, and the backbone does not receive any gradient.

A linear head $W_s \in \mathbb{R}^{768 \times 2}$ projects each patch feature to a two-channel shear signal, in the spirit of the in-plane components of a planar deformation field:
\begin{equation}
\footnotesize
\mathbf{s} = \mathbf{F} W_s + b_s \in \mathbb{R}^{B \times 2 \times 14 \times 14 \times 2}.
\end{equation}
We do not constrain $W_s$ to compute any particular differential operator, and it is learned end-to-end through the classification loss. This is a deliberate choice. Imposing closed-form operators such as curl or divergence as fixed features degrades transfer, for reasons we analyze in Section~\ref{sec:exp:ablation}. Leaving the projection learnable lets the linear head recombine the raw shear statistics into whatever signal is discriminative for the task at hand.

Motion direction lives in the change of the shear field between frames, since a static contact does not carry directional information. We take the difference between the two time-bins,
\begin{equation}
\footnotesize
\Delta \mathbf{s} = \mathbf{s}_{:,1,:,:,:} - \mathbf{s}_{:,0,:,:,:} \in \mathbb{R}^{B \times 14 \times 14 \times 2}, \quad \Delta \mathbf{s} = (\Delta s_x, \Delta s_y),
\end{equation}
where $\Delta s_x$ and $\Delta s_y$ are the two per-patch components.

\subsection{Contact Mask}
\label{sec:method:mask}

Not every patch carries genuine contact; many encode background gel deformation or imaging artifacts, and on a real sensor these regions are noisy. A second linear head $W_m \in \mathbb{R}^{768 \times 1}$ predicts a soft contact confidence per patch through a sigmoid, averaged over the two time-bins:
\begin{equation}
\footnotesize
\label{eq:contact_mask}
\begin{gathered}
m_{b,t,i,j}
= \sigma\!\left(\mathbf{F}_{b,t,i,j} W_m + b_m\right),\\
\bar{m}_{b,i,j}
= \frac{1}{T'} \sum_{t=0}^{T'-1} m_{b,t,i,j},
\quad
\bar{m} \in \mathbb{R}^{B \times 14 \times 14}.
\end{gathered}
\end{equation}
The mask is used as a spatial weighting in the pooling step below, not as a hard gate that zeros out patches.

\subsection{Antisymmetric Region Pooling}
\label{sec:method:pool}

The differential shear field is summarized over seven spatial regions, each pooled with the contact mask as a normalized weight. For a scalar field $f \in \mathbb{R}^{B \times 14 \times 14}$, with $f \in \{\Delta s_x, \Delta s_y\}$, and a region $R \subseteq \{0,\dots,13\}^2$, we define the mass-normalized regional mean
\begin{equation}
\footnotesize
\mathrm{Pool}_R(f) \;=\;
\frac{\sum_{(i,j)\in R}\, f_{ij}\, \bar{m}_{ij}}
     {\sum_{(i,j)\in R}\, \bar{m}_{ij}\, +\, \varepsilon},
\quad \varepsilon = 10^{-6}.
\label{eq:mass_norm_pool}
\end{equation}
Normalizing by the regional contact mass, rather than multiplying $f$ by $\bar m$, means the pooled value reports the average shear over the patches that are actually in contact, independent of how large the contact region is. The seven regions are
\begin{equation}
\footnotesize
\mathcal{R} \;=\; \{\,
\mathrm{U},\;
\mathrm{Lo},\;
\mathrm{U}{-}\mathrm{Lo},\;
\mathrm{Le},\;
\mathrm{Ri},\;
\mathrm{Le}{-}\mathrm{Ri},\;
\mathrm{G}
\,\},
\label{eq:seven_regions}
\end{equation}
with upper $\mathrm{U}$ and lower $\mathrm{Lo}$ halves split at the row midline ($\mathrm{rows}\,0{:}7$ and $7{:}14$), left $\mathrm{Le}$ and right $\mathrm{Ri}$ halves split at the column midline, and $\mathrm{G}$ the full grid. The two differenced regions are the antisymmetric terms:
\begin{equation}
\footnotesize
\label{eq:antisym_pool}
\begin{aligned}
P_{\mathrm{U-Lo}}(f)
= P_{\mathrm{U}}(f) - P_{\mathrm{Lo}}(f),
\quad
P_{\mathrm{Le-Ri}}(f)
= P_{\mathrm{Le}}(f) - P_{\mathrm{Ri}}(f),
\end{aligned}
\end{equation}
where $P$ denotes the pooling operator. They change sign when the contact motion is mirrored, which is precisely the behaviour of rotation chirality: a clockwise and a counter-clockwise twist produce opposite-signed shear across each half-plane pair. The remaining symmetric regions retain the magnitude information that sliding direction relies on, so the same pool serves both tasks.

\subsection{Geometric Feature and Fusion}
\label{sec:method:fusion}

Concatenating the seven pooled values over the two shear components gives a $14$-dimensional descriptor, which a linear head $W_o \in \mathbb{R}^{14 \times 768}$ lifts to the backbone hidden width:
\begin{equation}
\footnotesize
\mathbf{r} =
\mathrm{Conc}\!\Big[
\{\mathrm{P}_R(\Delta s_x)\}_{R \in \mathcal{R}},\;
\{\mathrm{P}_R(\Delta s_y)\}_{R \in \mathcal{R}}
\Big],
\quad
\mathbf{g} = W_o\,\mathbf{r} + b_o.
\label{eq:geom_token}
\end{equation}
The geometric feature $\mathbf{g}$ is fused with the frozen global token by addition, and a single linear layer maps the result to the class logits:
\begin{equation}
\footnotesize
\mathbf{z} = \mathbf{c} + \mathbf{g},
\quad
\hat{\mathbf{y}} = W_{\mathrm{cls}}\,\mathbf{z} + b_{\mathrm{cls}},
\label{eq:fusion}
\end{equation}
with $W_{\mathrm{cls}} \in \mathbb{R}^{768 \times C}$ and $C \in \{2, 4\}$ for rotation and sliding. Additive fusion keeps the probe input at the backbone width and lets the geometric feature act as a learned correction to the global representation rather than a separate channel. The three projections together, $W_s$ ($1{,}538$), $W_m$ ($769$), and $W_o$ ($11{,}520$), contribute $13{,}827$ trainable parameters, with the linear head this is the entire trainable budget on top of the frozen backbone. The module is implemented as the Differentiable Physics Encoder class. In the configuration used here its differential operators are disabled, so it runs as the purely geometric encoder described above.

\section{Experiments}
\label{sec:experiments}

We evaluate GeoTLM on two tactile motion-recognition tasks, rotation and sliding, across a simulated source domain and a real-sensor target domain. Our experiments are designed to answer three questions. First, does the geometric encoder improve discrimination of contact motion over strong tactile representation backbones under a controlled probing protocol? Second, does any such improvement survive the shift to a previously unseen physical sensor, and does it generalize to object instances held out from training? Third, which parts of the encoder are responsible for the gain, and does their relative importance change between clean simulation and noisy real data? We address these in turn after describing the setup.

\subsection{Experimental Setup}
\label{sec:exp:setup}

\noindent\textbf{Datasets.}
We use two datasets. ToucHD-Sim is the simulated subset of the AnyTouch~2 release~\cite{feng2026anytouch}, a pool of $113$ objects rendered under an optical tactile sensor model, which we split by motion type into a two-class rotation task ($2{,}034$ trials) and a four-class sliding task ($4{,}014$ trials, labelled left / right / up / down). TactileTracking is the real contact dataset released with NormalFlow~\cite{huang2024normalflow}, recorded on a physical GelSight Mini over $12$ everyday objects, with image statistics, gel deformation, and calibration that differ from the simulated source. Following the pairwise formulation of the source, each TactileTracking sample is a frame pair $(I_a, I_b)$ with a motion label. Rotation has $4{,}088$ pairs and sliding $6{,}400$ pairs.

\noindent\textbf{Evaluation protocols.}
On ToucHD-Sim we use five-fold object-disjoint cross-validation, so no object is shared between training and evaluation within a fold. On TactileTracking we report two protocols. Within-distribution is five-fold object-disjoint cross-validation, which we read as a fitting measure rather than a generalization claim. The novel-object protocol is leave-one-object-out (LOOO): each of the twelve objects is held out in turn as the sole evaluation object while the other eleven are used for training, so every object is tested once on a model that has never seen it. LOOO is the strict test of instance-level generalization and is our primary evidence.

\noindent\textbf{Implementation details.}
To isolate the representation from the classifier, all learning-based methods are evaluated identically: the encoder is frozen as a feature extractor, features are standardized per fold with statistics fit on the training partition alone, and a single linear layer is trained on top as the probe. For our method the standardization is applied to the frozen AnyTouch2 tokens before they enter DGR, since the geometric representation is recomputed per input. We train with AdamW (learning rate $10^{-3}$ for the head, $3\times10^{-4}$ for DGR), weight decay $0.01$, batch size $32$, for $30$ epochs, keeping the best-validation checkpoint. The seed is $42$ offset by the fold index. We compare against six backbones probed under this protocol: Sparsh~\cite{higuera2024sparsh}, InternViT-300M~\cite{chen2024internvl}, T3~\cite{zhao2024transferable}, AnyTouch~\cite{feng2025anytouch}, UniT~\cite{xu2025unit}, and AnyTouch2, which is our backbone stripped of the geometric encoder and serves as the controlled ablation. UniT is a caveated lower bound, as its markered pretraining is mismatched with our markerless data. For rotation we also include Curl + LR, a logistic-regression classifier on handcrafted shear-curl features, as a classical reference. We report mean and standard deviation of overall accuracy over folds, with per-class recall.

\vspace{-10pt}
\subsection{Rotation Tasks}
\label{sec:exp:rotation}

\begin{table*}[t]
  \centering
  \caption{Two-class rotation accuracy (clockwise / counter-clockwise) across simulated, real, and novel-object settings. $\ddagger$~UniT uses markered pretraining, mismatched with our markerless data.}
  \label{tab:rotation-compare}
  \footnotesize
  \setlength{\tabcolsep}{4pt}
  \resizebox{\linewidth}{!}{
  \begin{tabular}{lccccccccc}
    \toprule
      & \multicolumn{3}{c}{\textbf{ToucHD-Sim~\cite{feng2026anytouch}}}
      & \multicolumn{3}{c}{\textbf{TactileTracking~\cite{huang2024normalflow}}}
      & \multicolumn{3}{c}{\textbf{TactileTracking~\cite{huang2024normalflow} (novel-object)}} \\
    \cmidrule(lr){2-4} \cmidrule(lr){5-7} \cmidrule(lr){8-10}
      \textbf{Method}
      & \textbf{Overall} & \textbf{CW} & \textbf{CCW}
      & \textbf{Overall} & \textbf{CW} & \textbf{CCW}
      & \textbf{Overall} & \textbf{CW} & \textbf{CCW} \\
    \midrule
      Curl + LR
        & $0.890 \pm 0.237$ & $0.876$ & $0.904$
        & $0.742 \pm 0.072$ & $0.719$ & $0.765$
        & -- & -- & -- \\
      Sparsh V-JEPA~\cite{higuera2024sparsh}
        & $0.710 \pm 0.017$ & $0.731$ & $0.690$
        & $0.566 \pm 0.033$ & $0.516$ & $0.615$
        & $0.531 \pm 0.038$ & $0.510$ & $0.544$ \\
      InternViT-300M~\cite{chen2024internvl}
        & $0.823 \pm 0.023$ & $0.825$ & $0.822$
        & $0.552 \pm 0.045$ & $0.473$ & $0.613$
        & $0.536 \pm 0.037$ & $0.486$ & $0.581$ \\
      T3~\cite{zhao2024transferable}
        & $0.769 \pm 0.043$ & $0.749$ & $0.789$
        & $0.563 \pm 0.018$ & $0.497$ & $0.627$
        & $0.535 \pm 0.043$ & $0.533$ & $0.524$ \\
      AnyTouch (v1)~\cite{feng2025anytouch}
        & $0.733 \pm 0.027$ & $0.723$ & $0.743$
        & $0.550 \pm 0.057$ & $0.614$ & $0.488$
        & $0.513 \pm 0.062$ & $0.554$ & $0.483$ \\
      UniT\textsuperscript{$\ddagger$}~\cite{xu2025unit}
        & $0.685 \pm 0.035$ & $0.658$ & $0.711$
        & $0.550 \pm 0.042$ & $0.516$ & $0.581$
        & $0.515 \pm 0.044$ & $0.534$ & $0.506$ \\
      AnyTouch~2~\cite{feng2026anytouch}
        & $0.876 \pm 0.017$ & $0.872$ & $0.879$
        & $0.547 \pm 0.024$ & $0.538$ & $0.548$
        & $0.513 \pm 0.033$ & $0.440$ & $0.599$ \\
    \midrule
    \rowcolor{black!10}
      \textbf{GeoTLM (ours)}
        & $0.952 \pm 0.016$ & $0.953$ & $0.951$
        & $0.864 \pm 0.067$ & $0.856$ & $0.873$
        & $0.659 \pm 0.083$ & $0.692$ & $0.627$ \\
    \bottomrule
  \end{tabular}
  }
  \vspace{-10pt}
\end{table*}

Table~\ref{tab:rotation-compare} reports rotation accuracy across the three settings. On the simulated source, GeoTLM reaches $0.952$ overall against $0.876$ for the same backbone without the geometric encoder, and the prior backbones trail further behind. The simulated numbers are high across the board because the rendered contact images are clean, and they should be read as an upper reference rather than the operating point of interest.

The real-sensor results are where the encoders separate. Under within-distribution cross-validation GeoTLM reaches $0.864$, but we caution against reading this as a generalization figure: because the standardization and linear head are fit per fold and the validation objects partly recur across folds, this number reflects how well the representation fits this distribution, not how it transfers to a new instance. The table footnote makes this distinction explicit.

\begin{figure}[t]
\centering
\vspace{-25pt}
\includegraphics[width=\linewidth]{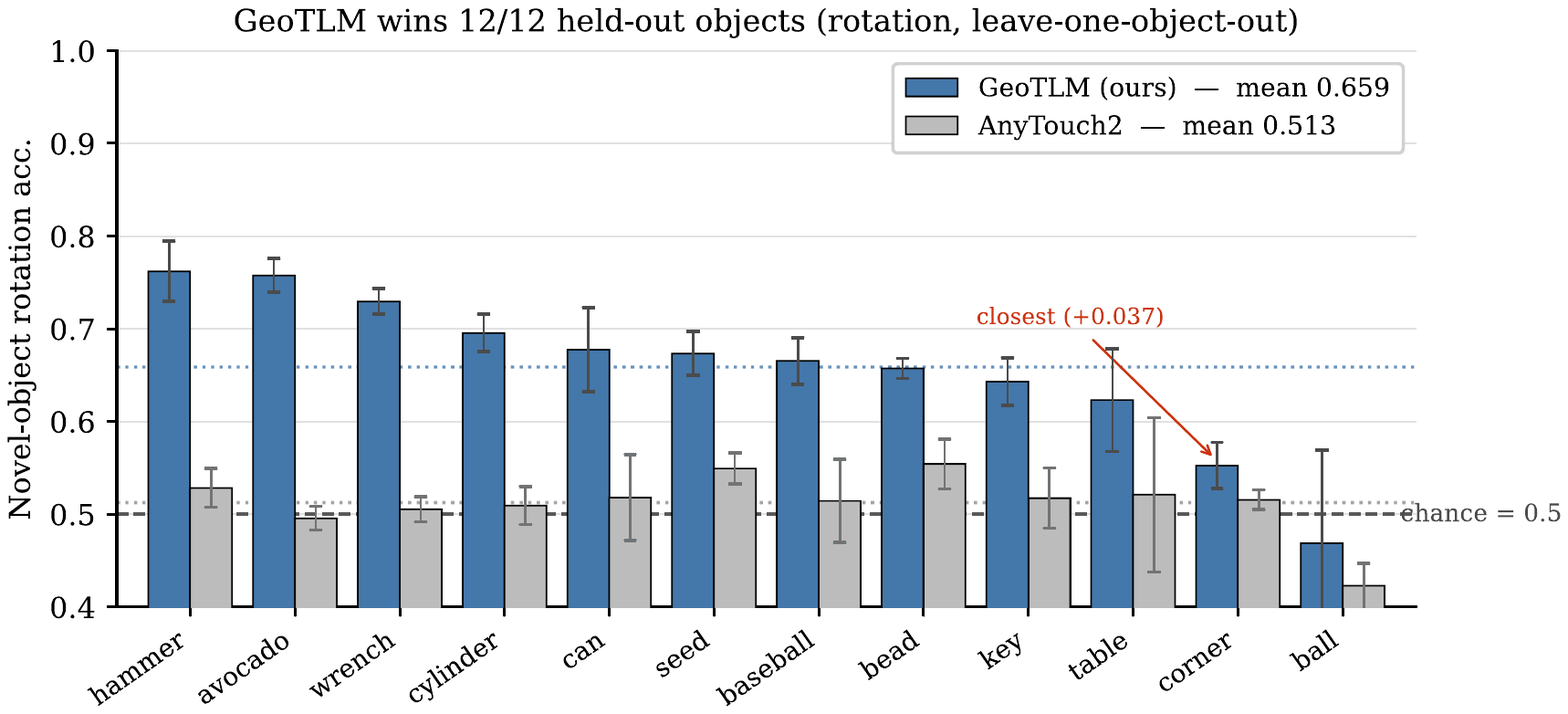}
\vspace{-48pt}
\caption{Per-object accuracy on the novel-object (leave-one-object-out) rotation task, comparing GeoTLM with AnyTouch~2 without the geometric encoder.}
\label{fig:looo_per_object}
\vspace{-12pt}
\end{figure}

The novel-object protocol is the decisive comparison. Under leave-one-object-out evaluation, where each test object is unseen during training, GeoTLM attains $0.659 \pm 0.083$ overall, against $0.513 \pm 0.034$ for the same backbone without DGR and $0.536$ for the strongest prior encoder (InternViT). This is an improvement of $+14.6$ percentage points over the DGR-ablated baseline and $+12.3$ points over the best prior representation. The gap is not driven by a few easy objects: DGR improves over the controlled baseline on all twelve held-out objects, with the smallest margin on the geometrically symmetric corner object ($+0.037$) and the two methods closest, though still separated, on the rotationally near-symmetric ball. Figure~\ref{fig:looo_per_object} shows the per-object breakdown.

\begin{figure}[t]
\centering
\includegraphics[width=\linewidth]{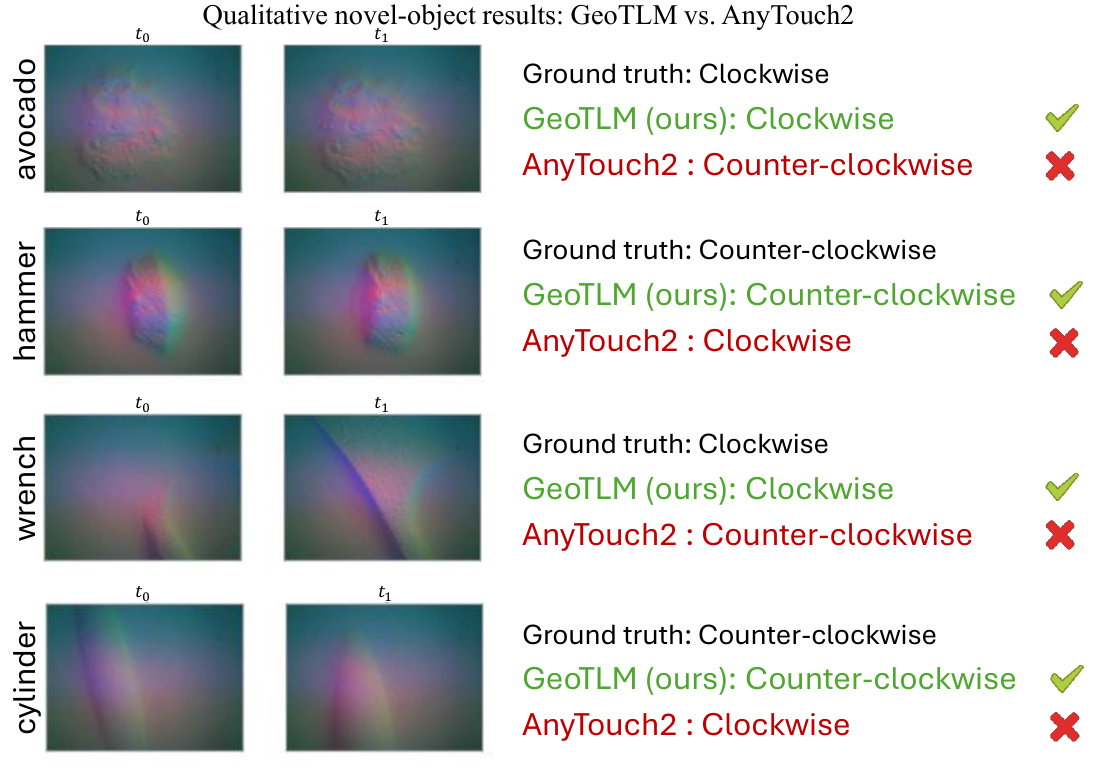}
\caption{Qualitative novel-object rotation results comparing GeoTLM with the DGR-ablated backbone. The two frames are temporally adjacent, making the rotation direction difficult to identify visually}
\label{fig:qualitative}
\vspace{-15pt}
\end{figure}

The per-class structure explains why the ablated baseline fails. Without the geometric encoder the probe collapses toward a single rotation direction on most held-out objects, leaving its balanced accuracy near the $0.50$ chance level even when its raw accuracy is slightly above it. GeoTLM keeps the two directions balanced ($0.69 / 0.63$ clockwise / counter-clockwise on the novel-object split), which indicates it supplies a chirality signal the frozen backbone does not expose on its own. A classifier given only the backbone features cannot recover this distinction across object instances. The antisymmetric region statistics computed by DGR make it available. Figure~\ref{fig:qualitative} shows representative held-out cases.

\subsection{Sliding Tasks}
\label{sec:exp:sliding}

\begin{table*}[t]
  \centering
  \caption{Four-class sliding direction accuracy (left / right / up / down; chance $=0.25$) across simulated, real, and novel-object settings. $\ddagger$~UniT uses markered pretraining, mismatched with our markerless data.}
  \label{tab:sliding-compare}
  \scriptsize
  \setlength{\tabcolsep}{3pt}
  \begin{tabular}{lccccccccccccccc}
    \toprule
      & \multicolumn{5}{c}{\textbf{ToucHD-Sim~\cite{feng2026anytouch}}}
      & \multicolumn{5}{c}{\textbf{TactileTracking~\cite{huang2024normalflow}}}
      & \multicolumn{5}{c}{\textbf{TactileTracking~\cite{huang2024normalflow} (novel-object)}} \\
    \cmidrule(lr){2-6} \cmidrule(lr){7-11} \cmidrule(lr){12-16}
      \textbf{Method}
      & \textbf{Overall} & \textbf{L} & \textbf{R} & \textbf{U} & \textbf{D}
      & \textbf{Overall} & \textbf{L} & \textbf{R} & \textbf{U} & \textbf{D}
      & \textbf{Overall} & \textbf{L} & \textbf{R} & \textbf{U} & \textbf{D} \\
    \midrule
      Sparsh V-JEPA~\cite{higuera2024sparsh}
        & $0.710 \pm 0.036$ & $0.715$ & $0.729$ & $0.686$ & $0.709$
        & $0.384 \pm 0.046$ & $0.381$ & $0.131$ & $0.534$ & $0.333$
        & $0.365 \pm 0.037$ & $0.376$ & $0.202$ & $0.480$ & $0.348$ \\
      InternViT-300M~\cite{chen2024internvl}
        & $0.907 \pm 0.034$ & $0.919$ & $0.875$ & $0.929$ & $0.904$
        & $0.338 \pm 0.016$ & $0.495$ & $0.207$ & $0.313$ & $0.335$
        & $0.291 \pm 0.058$ & $0.403$ & $0.234$ & $0.281$ & $0.251$ \\
      T3~\cite{zhao2024transferable}
        & $0.857 \pm 0.023$ & $0.844$ & $0.880$ & $0.849$ & $0.857$
        & $0.383 \pm 0.033$ & $0.350$ & $0.229$ & $0.520$ & $0.347$
        & $0.325 \pm 0.048$ & $0.342$ & $0.226$ & $0.374$ & $0.321$ \\
      AnyTouch (v1)~\cite{feng2025anytouch}
        & $0.809 \pm 0.006$ & $0.806$ & $0.816$ & $0.794$ & $0.823$
        & $0.334 \pm 0.039$ & $0.303$ & $0.103$ & $0.568$ & $0.219$
        & $0.286 \pm 0.036$ & $0.285$ & $0.204$ & $0.367$ & $0.251$ \\
      UniT\textsuperscript{$\ddagger$}~\cite{xu2025unit}
        & $0.691 \pm 0.028$ & $0.666$ & $0.672$ & $0.729$ & $0.697$
        & $0.346 \pm 0.013$ & $0.380$ & $0.262$ & $0.377$ & $0.331$
        & $0.317 \pm 0.026$ & $0.360$ & $0.256$ & $0.347$ & $0.245$ \\
      AnyTouch~2~\cite{feng2026anytouch}
        & $0.967 \pm 0.014$ & $0.957$ & $0.960$ & $0.975$ & $0.974$
        & $0.363 \pm 0.048$ & $0.313$ & $0.192$ & $0.614$ & $0.256$
        & $0.328 \pm 0.036$ & $0.298$ & $0.197$ & $0.438$ & $0.307$ \\
    \midrule
    \rowcolor{black!10}
      \textbf{GeoTLM (ours)}
        & $0.989 \pm 0.008$ & $0.991$ & $0.985$ & $0.988$ & $0.990$
        & $0.525 \pm 0.088$ & $0.500$ & $0.316$ & $0.631$ & $0.577$
        & $0.396 \pm 0.051$ & $0.347$ & $0.302$ & $0.458$ & $0.431$ \\
    \bottomrule
  \end{tabular}
  \vspace{-5pt}
  
\end{table*}

Table~\ref{tab:sliding-compare} reports the four-class sliding task, where the chance level is $0.25$. On the simulated source GeoTLM again leads, at $0.989$ against $0.967$ for the ablated backbone. On the real GelSight Mini data the geometric encoder lifts overall accuracy from $0.363$ to $0.525$, a gain of $+16.2$ points, while every prior backbone remains close to the $0.34$–$0.38$ band. The per-class recalls show that the prior backbones primarily exhibit class collapse rather than uniformly distributed errors. Their accuracy concentrates on the vertical directions and the horizontal directions are largely suppressed. GeoTLM recovers a more even distribution and is strongest on the up and down classes ($0.63 / 0.58$), with the rightward class the hardest for every method.
We also report a novel-object sliding column in Table~\ref{tab:sliding-compare}. This setting remains challenging since sliding direction is only weakly recoverable from GelSight contact image and the margins between methods are narrow. Still, GeoTLM attains the highest novel-object accuracy ($0.396$), ahead of the strongest baseline Sparsh ($0.365$), which we have read as a step forward on a hard problem.

\vspace{-5pt}
\subsection{Ablation and Analysis}
\label{sec:exp:ablation}

\begin{table*}[t]
  \centering
  \caption{Ablation of the two DGR components (region pool, contact mask) on two-class rotation.}
  \label{tab:ablation}
  \footnotesize
  \setlength{\tabcolsep}{6pt}
  \resizebox{\linewidth}{!}{
  \begin{tabular}{lcccccccc}
    \toprule
      & & 
      & \multicolumn{3}{c}{\textbf{ToucHD-Sim~\cite{feng2026anytouch}}}
      & \multicolumn{3}{c}{\textbf{TactileTracking~\cite{huang2024normalflow}}} \\
    \cmidrule(lr){4-6} \cmidrule(lr){7-9}
      \textbf{Variant}
      & \textbf{Region} & \textbf{Mask}
      & \textbf{Overall} & \textbf{CW} & \textbf{CCW}
      & \textbf{Overall} & \textbf{CW} & \textbf{CCW} \\
    \midrule
      AnyTouch~2~\cite{feng2026anytouch}
        & \ding{55} & \ding{55}
        & $0.876 \pm 0.017$ & $0.872$ & $0.879$
        & $0.547 \pm 0.024$ & $0.538$ & $0.548$ \\
      DGR $-$ contact mask
        & \ding{51} & \ding{55}
        & $0.946 \pm 0.014$ & $0.939$ & $0.952$
        & $0.768 \pm 0.064$ & $0.772$ & $0.765$ \\
      DGR $-$ region pool
        & \ding{55} & \ding{51}
        & $0.918 \pm 0.010$ & $0.923$ & $0.912$
        & $0.781 \pm 0.096$ & $0.783$ & $0.781$ \\
    \midrule
    \rowcolor{black!10}
      \textbf{GeoTLM (ours)}
        & \ding{51} & \ding{51}
        & $0.952 \pm 0.016$ & $0.953$ & $0.951$
        & $0.864 \pm 0.067$ & $0.856$ & $0.873$ \\
    \bottomrule
  \end{tabular}
  }
  \vspace{-10pt}
\end{table*}

DGR combines two mechanisms on top of the frozen backbone features: a seven-region antisymmetric pool over the shear field, and a contact mask that weights each region by its contact mass. Table~\ref{tab:ablation} removes each in turn on the two-class rotation task, on both the simulated source and the real-sensor cross-validation split. The endpoints coincide with the corresponding rows of the main rotation table, confirming that the ablation and main comparison share one protocol.

The two domains show different ablation patterns, and the contrast is itself informative. On clean simulation the region pool carries almost the entire effect: adding it alone raises accuracy from $0.876$ to $0.946$, while the contact mask adds little once the region pool is present (its marginal contribution is $+0.006$). The full encoder improves over the ablated backbone by $+0.076$ here. On the real-sensor data the picture changes. The total improvement is $+0.317$, roughly four times larger than in simulation, and the two components are now both necessary and complementary: the region pool contributes $+0.083$ at the margin, the contact mask $+0.096$, and the mask alone ($0.781$) is no weaker than the region pool alone ($0.768$). The reading we draw is that the contact mask becomes important under real-sensor noise. In simulation the non-contact regions of the gel are clean and carry no spurious signal, so down-weighting them changes little. On a physical sensor the non-contact regions are noisy, and suppressing them is what allows the region statistics to remain discriminative. The component that appears redundant on clean data is the one that protects the representation on real data.

The frame-count ablation, which controls for the number of input frames rather than for the encoder, is reported in Table~\ref{tab:frame-ablation}. It shows that at an equal two-frame input on the same backbone the geometric encoder still accounts for the majority of the gain.

\begin{table}[t]
  \centering
  \caption{Effect of input frame count versus the DGR module on ToucHD-Sim rotation.}
  \label{tab:frame-ablation}
  \footnotesize
  \resizebox{\linewidth}{!}{
  \begin{tabular}{lc}
    \toprule
    \textbf{Variant (ToucHD-Sim rotation)} & \textbf{Overall} \\
    \midrule
    AnyTouch~2, 2-frame & $0.822 \pm 0.022$ \\
    AnyTouch~2, 4-frame & $0.876 \pm 0.017$ \\
    GeoTLM (ours), 2-frame          & $0.902 \pm 0.025$ \\
    \rowcolor{black!10}
    \textbf{GeoTLM (ours)}, 4-frame          & $0.952 \pm 0.016$ \\
    \bottomrule
  \end{tabular}
  }
  \vspace{-15pt}
\end{table}

\vspace{-5pt}
\section{Conclusion}
\label{sec:conclusion}

We presented GeoTLM, a compact differentiable geometric encoder that recovers contact-motion direction from a frozen tactile backbone. DGR projects patch tokens to a learned shear field, differences it across frames, and pools it over antisymmetric, contact-mass-weighted regions whose sign structure mirrors the chirality of in-plane rotation. With the backbone fixed and only about 14k trained parameters, it turns representations a linear probe cannot read into ones that support reliable rotation and sliding discrimination. On a physical GelSight Mini and objects unseen during training, DGR improves novel-object rotation by $14.6$ points over the same backbone without it and by $12.3$ points over the strongest prior representation, winning on all twelve held-out objects. The gain is substantially larger on real data than in simulation, highlighting the practical value of the geometric prior under sensor noise and domain shift.
Several directions remain open. The current encoder targets planar rotation and sliding, and extending the antisymmetric pooling to richer contact dynamics such as three-dimensional twist or multi-contact events is a natural next step. Finally, coupling DGR with a language head for closed-loop manipulation, where motion-direction perception feeds task-level reasoning, is a promising path toward contact-rich robotic control.


 

\bibliographystyle{IEEEtran}
\bibliography{main}


\newpage

 




\vfill

\end{document}